\documentclass[letterpaper]{article} 
\usepackage{aaai2026}  
\usepackage{times}  
\usepackage{helvet}  
\usepackage{courier}  
\usepackage[hyphens]{url}  
\usepackage{graphicx} 
\usepackage{natbib}  
\usepackage{caption} 
\usepackage{algorithm}
\usepackage{algorithmic}
\usepackage{amsmath}
\usepackage{booktabs}
\usepackage{amsfonts}

\usepackage{newfloat}
\usepackage{listings}
\DeclareCaptionStyle{ruled}{labelfont=normalfont,labelsep=colon,strut=off} 
\floatstyle{ruled}
\newfloat{listing}{tb}{lst}{}
\floatname{listing}{Listing}
\title{Scientifically-Interpretable Reasoning Network (ScIReN): \\ Discovering Hidden Relationships in the Carbon Cycle and Beyond}
\author{
    Joshua Fan\equalcontrib\textsuperscript{\rm 1},
    Haodi Xu\equalcontrib\textsuperscript{\rm 2},
    Feng Tao\equalcontrib\textsuperscript{\rm 3}\thanks{Now at Department of Informatics and Intelligent Systems, Institute of Energy and the Environment, The Pennsylvania State University.},
    Md Nasim\textsuperscript{\rm 1},
    Marc Grimson\textsuperscript{\rm 1},
    Yiqi Luo\textsuperscript{\rm 2},
    Carla P. Gomes\textsuperscript{\rm 1}
}
\affiliations{
    \textsuperscript{\rm 1}Cornell University, Department of Computer Science \\
    \textsuperscript{\rm 2}Cornell University, School of Integrative Plant Science, Soil and Crop Sciences Section \\
    \textsuperscript{\rm 3}Cornell University, Department of Ecology and Evolutionary Biology \\

    \{jyf6,hx293,feng.tao,mn637,mg2422,yl2735\}@cornell.edu, gomes@cs.cornell.edu
}

\usepackage{bibentry}

\begin{document}

\maketitle



\begin{abstract}
    Soils have potential to mitigate climate change by sequestering carbon from the atmosphere, but the soil carbon cycle remains poorly understood. Scientists have developed \emph{process-based models} of the soil carbon cycle based on existing knowledge, but they contain numerous unknown parameters and often fit observations poorly. On the other hand, neural networks can learn patterns from data, but do not respect known scientific laws, and are too opaque to reveal novel scientific relationships. We thus propose \emph{Scientifically-Interpretable Reasoning Network (ScIReN)}, a fully-transparent framework that combines interpretable neural and process-based reasoning. An interpretable encoder predicts \textit{scientifically-meaningful latent parameters}, which are then passed through a differentiable process-based decoder to predict labeled output variables. While the process-based decoder enforces existing scientific knowledge, the encoder leverages Kolmogorov-Arnold networks (KANs) to reveal interpretable relationships between input features and latent parameters, using novel smoothness penalties to balance expressivity and simplicity. ScIReN also introduces a novel hard-sigmoid constraint layer to restrict latent parameters into prior ranges while maintaining interpretability. We apply ScIReN on two tasks: simulating the flow of organic carbon through soils, and modeling ecosystem respiration from plants. On both tasks, ScIReN outperforms or matches black-box models in predictive accuracy, while greatly improving scientific interpretability -- it can infer latent scientific mechanisms and their relationships with input features.
    

\end{abstract}

\begin{links}
    \link{Code}{https://github.com/gomes-lab/ScIReN}
\end{links}

\section{Introduction}

Climate change poses major challenges to humanity, such as sea level rise, damages to agricultural production, and increased natural disasters \cite{lee2023ipcc}. Soils have the potential to mitigate climate change by sequestering carbon from the atmosphere \cite{lal2015carbon}, as they store more carbon than plants and the atmosphere combined \cite{jobbagy2000vertical}. Yet it is difficult to understand how carbon flows through the soil, leading to high uncertainties in future climate projections \cite{luo2016toward}. 

Neural networks can predict the  amount of organic carbon stored in the soil at each location \cite{tan2024importance}. However, neural networks can produce predictions that violate existing scientific knowledge, such as mass conservation \cite{karpatne2022knowledge}. Also, scientists want to \emph{understand} the biogeochemical processes governing the soil carbon cycle -- which control how much carbon can be stored in the soil and how long it will remain sequestered there. However, the black-box nature of neural networks makes it challenging to translate their predictive ability into novel scientific insights \cite{marcinkevivcs2023interpretable}. 

To gain insight into these biogeochemical processes, ecologists use  domain expertise to develop process-based models that simulate the soil carbon cycle \cite{luo2022land}. 
These models are highly sophisticated, and may contain hundreds of pools representing different types of carbon, and matrices specifying flux rates between each pair of pools \cite{luo2022matrix, huang2018matrix, lu2020full}. However, these models contain many unknown parameters, which traditionally must be tuned by human experts through an inefficient trial-and-error process \cite{luo2020model}. These models often cannot fit observed data well, especially in cross-scale predictions, due to a poor understanding of the relationships between environmental conditions and parameter values. This is a key bottleneck in using process-based models for verifiable predictions on the global soil carbon cycle and its response to climate change.

A few prior works in hydrology \cite{tsai2021calibration,feng2022differentiable}, phenology \cite{van2025hybrid}, ecosystem modeling \cite{reichstein2022combining}, and soil biogeochemistry \cite{xu2025biogeochemistry} have proposed implementing process-based models in a differentiable way so that poorly-understood processes can be optimized by backpropagation or replaced with neural networks. While these approaches integrate scientific knowledge with deep learning, the neural network components remain too opaque for scientists to understand, which leaves the relationships between the latent parameters and input features unclear.

To address these limitations, we propose \emph{Scientifically-Interpretable Reasoning Network (ScIReN)}, an end-to-end differentiable framework that embeds scientific process-based models into a \emph{fully-transparent} neural model, creating a system that respects scientific knowledge, learns from data in an end-to-end manner, and can discover new scientific relationships. ScIReN contains three main components. First, a learnable encoder takes in environmental features for a given location, and predicts scientifically-meaningful latent parameters. We use an \emph{interpretable model} such as a sparse Kolmogorov-Arnold network \cite{liu2024kan} to make this portion fully transparent, allowing scientists to understand the relationships between latent parameters and input features. Second, a hard-sigmoid constraint layer projects these parameters into a physically-plausible range set by prior knowledge. Finally, a process-based decoder uses these predicted parameters to simulate the flow of carbon through the soil based on scientific knowledge, and finally predicts the output variables (e.g. amount of organic carbon at each soil depth). We compare our predictions with ground-truth labels, and backpropagate to train the entire system. 

\textbf{Our main contributions are:} \textbf{(1)} We propose ScIReN, a \emph{fully-interpretable} 
framework for combining scientific reasoning (in the form of process-based models) with data-driven learning. Unlike prior work, our model can infer \emph{unobserved physical processes} 
and \emph{their relationships to input variables} in a fully-transparent way. \textbf{(2)} We balance smoothness and expressivity in the learned functional relationships by using B-splines with novel smoothness penalties. \textbf{(3)} We propose a novel hard-sigmoid constraint layer to constrain the scientific parameters to fall within physically-plausible ranges. \textbf{(4)} We validate ScIReN on two scientific domains. First, in ecosystem respiration, ScIReN discovers the correct relationships between latent parameters and input features, while other methods do not. It also improves out-of-distribution extrapolation compared to existing methods. Second, we test ScIReN on a more complex task of modeling the soil carbon cycle through 140 soil pools at each location. On a synthetic dataset, ScIReN is able to predict \emph{unlabeled} biogeochemical parameters and accurately retrieve their relationships with environmental features. With real data, ScIReN simulates observed soil carbon amounts with accuracy comparable to black-box methods while being completely interpretable. We hope our work inspires other researchers to apply and extend ScIReN across diverse scientific tasks, advancing AI’s capacity for interpretable scientific discovery.




\section{Related Work}

\paragraph{Knowledge-Guided Machine Learning.} There is a rich history of incorporating prior knowledge into neural networks by modifying the loss function, pretraining procedure, or model architecture \cite{karpatne2024knowledge, willard2022integrating}. A common approach is to add a loss term that penalizes when physical laws are violated \cite{daw2022physics, beucler2019achieving, jia2019physics}. For example, the density of water is known to increase monotonically with depth; thus, \citet{daw2022physics} and \citet{jia2019physics} use a ``monotonicity loss'' to penalize violations of this. Physics-informed neural networks assume that the governing equation of a system is known, and penalize when the model predictions or gradients violate this equation \cite{raissi2019physics}, 
but are hard to train \cite{krishnapriyan2021characterizing}.
Process-based models can generate synthetic data to pretrain the network \cite{jia2021physics,liu2024knowledge}. While these approaches use prior knowledge to guide the model, they cannot guarantee that physical constraints will be satisfied \cite{willard2022integrating}, and cannot easily provide new insights into physical processes.

One can also design model architectures to encode prior knowledge. Convolutional neural networks encode inductive biases such as translation equivariance and locality into the model architecture \cite{lecun1995convolutional}. In lake temperature modeling, \citet{daw2020physics} design an LSTM variant that produces monotonically-increasing intermediate variables by design. In agriculture, \citet{liu2024knowledge} design a hierarchical neural network that incorporates causal relations between different variables. However, it is difficult to design a new architecture for every problem. 

\paragraph{Combining reasoning and learning.} A few works reason about constraints and prior knowledge within the network itself. Deep Reasoning Networks use entropy-based losses to encourage the latent space to be interpretable and satisfy constraints (e.g. in Sudoku, each row must contain exactly one of each number) \cite{chen2020deep}. This approach was used to solve the phase-mapping problem in materials discovery, where the constraints are thermodynamic rules \cite{chen2021automating}. CLR-DRNets enhanced the reasoning process using a modified LSTM, and used curriculum learning to improve trainability \cite{bai2021clr}.  Physically-informed Graph-based DRNets add a physical decoder that reconstructs X-ray diffraction patterns based on Bragg’s law \cite{min2023physically}.  
CS-SUNet encourages pixels with similar input features to have similar predictions, enabling weakly-supervised prediction 
\cite{fan2022monitoring}. While these models have an interpretable latent space, the bulk of the network remains uninterpretable.

\paragraph{Process-based models.} Scientists  develop \emph{process-based models} to simulate physical processes based on domain knowledge \cite{cuddington2013process}. These models consist of mathematical equations that describe relationships between various variables. In soil science, \emph{pool-and-flux} models are common, where a matrix equation tracks the amount of carbon at each soil depth and matter type \cite{luo2022land}. Transition matrices encode the rate at which carbon is transferred between pools, which are functions of soil and climate properties. Many scientific models can be unified under this matrix form \cite{huang2018matrix, luo2022matrix}. 

Unfortunately, despite their sophistication, these models have difficulty matching real observations, and have numerous unknown parameters that are traditionally set in an ad-hoc way. These unknown parameters need to vary across space and (sometimes) time, but are difficult to estimate \cite{luo2020model}. A state-of-the-art approach for setting these parameters is PRODA \cite{tao2020deep}. PRODA first runs Bayesian data assimilation at each location separately to find optimal biogeochemical parameters for each location. Then, a neural network is trained to predict these optimal parameters given environmental covariates. While this approach is effective, it is computationally expensive, and is not always robust since each location's parameters are estimated with only a few observations.

\paragraph{Differentiable Process-Based Models.} A few works integrate process-based models and neural networks in an end-to-end differentiable framework; this has been called \emph{differentiable parameter learning} \cite{tsai2021calibration}, \emph{hybrid modeling} \cite{reichstein2022combining}, or \emph{differentiable process-based modeling} \cite{shen2023differentiable}. For example, \cite{van2025hybrid, reichstein2022combining, tsai2021calibration} used a process-based model as the main backbone, but replaced some poorly-understood components with neural networks. 
By implementing the process-based model in a differentiable way, the model could be trained end-to-end, and unknown components could be fit using data. 
\citet{xu2025biogeochemistry} scaled this approach up to a more complex soil carbon model with 21 unknown parameters and 140 carbon pools. However, the neural network component is still opaque, making it difficult for scientists to discover new relationships and insights. 


\paragraph{Interpretable ML.} A subfield of machine learning aims to interpret how neural networks make predictions \cite{molnar2025}. \emph{Post-hoc feature attribution methods} such as SHAP \cite{lundberg2017unified} or Integrated Gradients \cite{sundararajan2017axiomatic} estimate the impact of each feature on the model's prediction for a given example. \emph{Local surrogate models} such as LIME \cite{ribeiro2016model} fit an interpretable model that approximates the black-box model in a small region. \emph{Marginal effect plots} such as Partial Dependence Plots \cite{friedman2001greedy} or Accumulated Local Effects plots \cite{apley2020visualizing} plot how each feature affects the output on average. These methods are mere approximations of a black-box model, and can produce misleading explanations \cite{rudin2019stop}.

On the other hand, \emph{inherently-interpretable models} make the entire model transparent by design \cite{rudin2019stop}. In linear regression, coefficients precisely reveal how each input affects the prediction. Neural additive models \cite{agarwal2021neural} and Kolmogorov-Arnold Networks \cite{liu2024kan} provide greater expressivity while maintaining interpretability; we discuss these in the next section. These methods are typically applied in supervised settings; to the best of our knowledge, our work is the first to combine them  with scientific knowledge to predict \emph{unlabeled} variables.
\section{Methods}

\begin{figure*}
  \centering
\includegraphics[width=0.98\textwidth,keepaspectratio]{figures/ScIReN_Structure.png} 
  \caption{Overview of ScIReN. The encoder reveals interpretable functional relationships between environmental inputs (e.g. temperature) and latent scientific parameters (e.g. transfer rates between soil pools). A constraint layer forces latent parameters into a prior range, and the process-based decoder simulates the physical process with the given latent parameters.}
  \label{method_diagram}
\end{figure*} 

To combine scientific knowledge and data-driven learning into a fully-transparent model, ScIReN contains three main components. First, a learned encoder $f_{NN}$ (with learnable weights $\theta$) takes in input features $\mathbf{x} \in \mathbb{R}^D$ (e.g. soil/climate variables at a given location), and outputs unconstrained latent parameters $\tilde{\mathbf{p}} \in \mathbb{R}^P$: $\tilde{\mathbf{p}} = f_{NN}(\mathbf{x}; \theta)$. In ScIReN, $f_{NN}$ should be a fully-transparent model, such as a neural additive model or sparse Kolmogorov-Arnold network. The latent parameters are scientifically-meaningful variables that govern underlying physical processes, yet cannot be observed directly.  Secondly, a constraint layer projects the unconstrained parameters into scientifically-plausible ranges given by prior knowledge: $\mathbf{p} = \text{Proj}(\tilde{\mathbf{p}})$. Finally, the constrained parameters are passed through a \emph{fixed, deterministic} process-based decoder $g_{PBM}$, which simulates physical processes and predicts output variables: $\hat{y} = g_{PBM}(\mathbf{p})$. We compare this with the true label and backpropagate. The framework is summarized in Figure \ref{method_diagram}, and we elaborate on each component below.

\subsection{Encoder: Learned Interpretable Relationships}

We want to learn a function $f_{NN}$ mapping observed input features $\mathbf{x}$ to latent scientific parameters $\mathbf{p}$. In prior work, $f_{NN}$ is typically a fully-connected neural network \cite{xu2025biogeochemistry, van2025hybrid, reichstein2022combining, tsai2021calibration}. However, 
 scientists want to understand how biogeochemical parameters (e.g. transfer rates between pools) depend on input features (e.g. temperature); neural networks do not provide these insights.

Recently, a line of work has aimed to produce neural networks that are inherently interpretable while being expressive. Neural additive models (NAM) \cite{agarwal2021neural} model the output as the sum of single-variable functions of each input feature. Specifically, they learn a neural network $\phi_j : \mathbb{R} \rightarrow \mathbb{R}$ (with one input and one output) for each feature $x_j$, and sum contributions from each feature into the output:
\begin{equation}
    NAM(\mathbf{x}) = b + \sum_{j=1}^D \phi_j(x_j; \theta_j)
\end{equation}
where $(b, \{\theta_j\}_{j=1}^D)$ are learnable parameters trained via backpropagation. While NAM is quite expressive, it cannot model non-additive feature interactions, which are important in soil science \cite{dieleman2012simple}. 

To increase the expressivity of neural additive models, we can generate intermediate variables using neural additive model of the inputs, then further use a neural additive model on the intermediate variables to generate the output. Specifically, generate intermediate variables $\mathbf{z} = \{z_1, \dots, z_H\}$ as
\begin{equation}
z_h = NAM_h (\mathbf{x}) = b_h + \sum_{j=1}^D \phi_{h, j}(x_j), \quad \forall h \in [1, H]
\end{equation}
Now define each output variable $\tilde{p}_i$ as a neural additive model over the intermediate variables
\begin{align}
    \tilde{p}_i = NAM_i(\mathbf{z}) &= b_i + \sum_{h=1}^H \Phi_{i,h} (z_h) \\
    &= b_i' + \sum_{h=1}^H \Phi_{i,h} \left( \sum_{j=1}^D \phi_{h,j}(x_j) \right)
\end{align}
where all the bias terms are collected into $b_i'$. This is now the same form as a two-layer Kolmogorov-Arnold network (KAN, see equation 2.1 in \cite{liu2024kan}). By the Kolmogorov-Arnold theorem, this multi-layer stack of neural additive models can approximate any multivariate continuous function \cite{liu2024kan}.

\begin{figure}
  \centering
  \includegraphics[width=0.49\columnwidth]{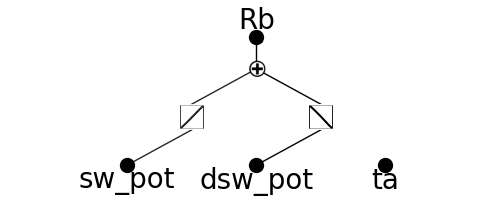}
  \includegraphics[width=0.49\columnwidth, trim={0 0.35cm 0 0, clip}]{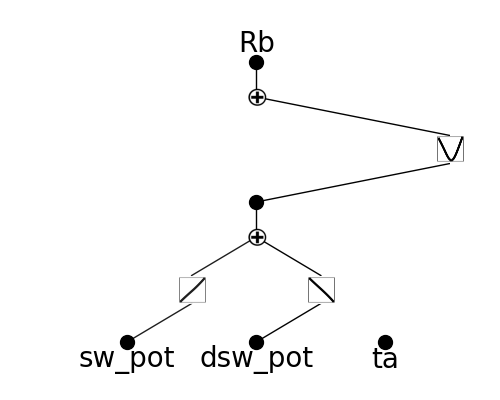}
  \caption{Learned encoder examples: 1-layer KAN (left) and 2-layer KAN (right)}
  \label{encoder}
\end{figure}

Two examples of learned encoders are shown in Figure \ref{encoder}. On the left, a 1-layer KAN models the latent parameter $Rb \approx w_1 \cdot sw\_pot + (-w_2) \cdot dsw\_pot$. One can immediately see how each feature influences $Rb$. On the right, a 2-layer KAN is needed to capture this nonlinear relationship, where an intermediate variable $Rb' \approx w_1 \cdot sw\_pot + (-w_2) \cdot dsw\_pot$, and then $Rb \approx |Rb'|$.

\subsection{Sparsity and Smoothness Regularization}

Kolmogorov-Arnold networks are still hard to interpret if each output depends on many inputs.
\citet{liu2024kan2} propose entropy regularization to sparsify the network. Specifically, we compute an importance score for each edge, as the mean absolute deviation of the output activations from the edge, weighted by their eventual contribution to the final output variables \cite{liu2024kan2}; see Appendix \ref{edge_score} for details. We denote this score for edge $(i \rightarrow j)$ in layer $l$ as $B_{i,j}^l$. We encourage the entropy of the edge importance distribution to be low (making the network choose a few important edges and push others towards zero). We also use a L1 penalty to further promote sparsity by shrinking the magnitudes of each edge's outputs towards zero: 
\begin{equation}
    e_{i,j}^l = \frac{B_{i,j}^l}{\sum_{i',j'} B_{i',j'}^l} \quad  \text{(make edge importances sum to 1)}
\end{equation}
\begin{equation}
    \mathcal{L}_{entropy} = -\sum_{l} \sum_{i,j} e_{i,j}^l \log e_{i,j}^l; \ \mathcal{L}_{L1} = -\sum_{l} \sum_{i,j} |B_{i,j}^l|
\end{equation}
The model can be viewed as using gradient descent to reason over possible scientific relationships, ultimately choosing a sparse set of relationships that fits the data.

Note that KANs parameterize the learnable edge activation functions $\Phi,\phi$ using B-splines instead of neural networks. B-splines represent a curve as a weighted sum of basis functions, each of which peaks at a different point on the x-axis; see \cite{eilers1996flexible} for details. We can apply a second-order difference penalty on the spline coefficients -- this encourages the coefficients to change in a linear way, making the function more linear \cite{eilers2010splines}. To our knowledge this has not been proposed with KANs before; this allows us to increase the number of basis functions (knots) and the function's expressivity while maintaining smoothness and preventing overfitting.
If $c_1 \dots c_G$ are B-spline coefficients, the penalty is
\begin{equation} 
\mathcal{L}_{smooth} = \sum_{i=1}^{G-2} ((c_{i+2} - c_{i+1}) - (c_{i+1} - c_i))^2
\end{equation}

\subsection{Linear Parameter Constraint Layer}

For some latent scientific parameters $p_i$, there is a known prior range $[p_i^{min}, p_i^{max}]$ based on prior knowledge and physical plausibility. \citet{xu2025biogeochemistry} applied a sigmoid function to the encoder output $\tilde{p}_i$ to force the predicted parameter into the prior range: $p_i = \sigma(\tilde{p}_i)$. However, this adds nonlinearity and harms interpretability. For example, suppose a parameter $p_i$ is actually a linear function of input variable $x_j$, $p_i = wx_j$. If the parameter was constrained using a sigmoid, the unconstrained encoder would have to learn $\tilde{p}_i = \sigma^{-1} (wx_j)$, so that after the sigmoid function the parameter becomes $p_i = \sigma(\sigma^{-1}(wx_j)) = w x_j$. The additional inverse sigmoid makes the function less interpretable.

\begin{figure}
  \centering
  \includegraphics[width=0.45\columnwidth]{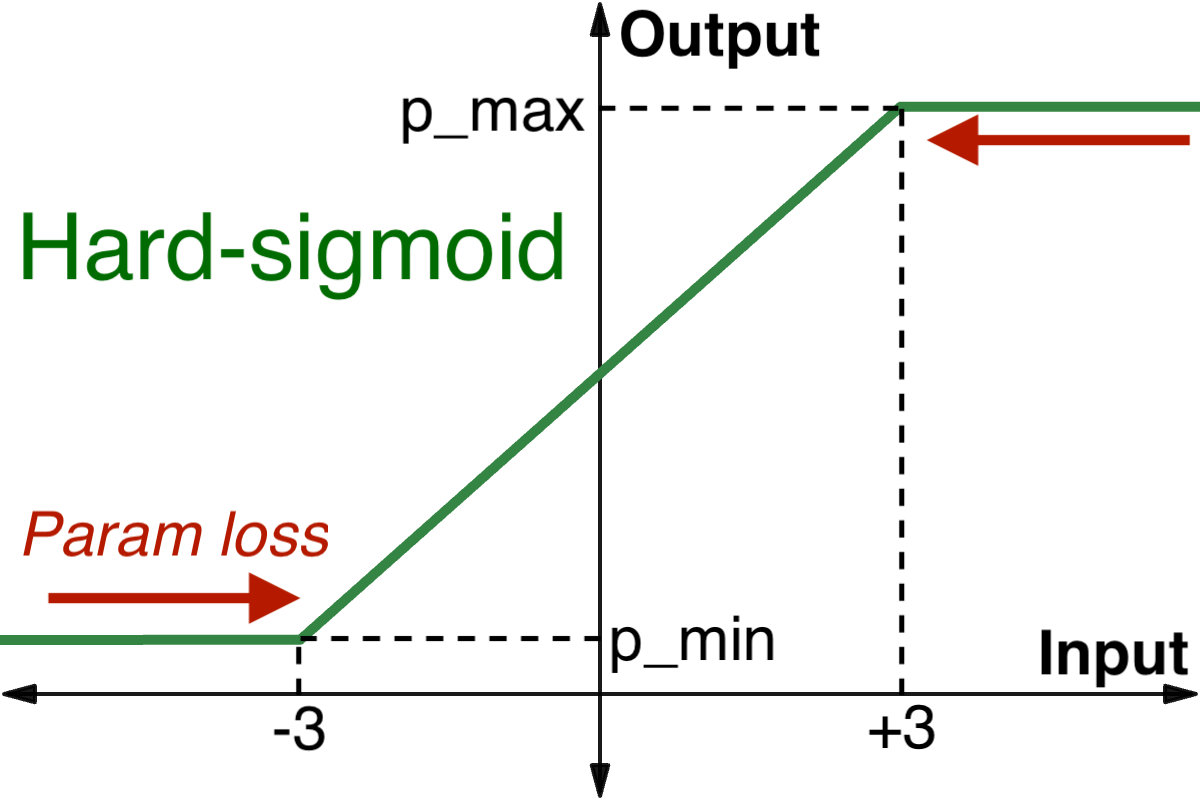}
  \hspace{0.08\columnwidth}
  \includegraphics[width=0.45\columnwidth]{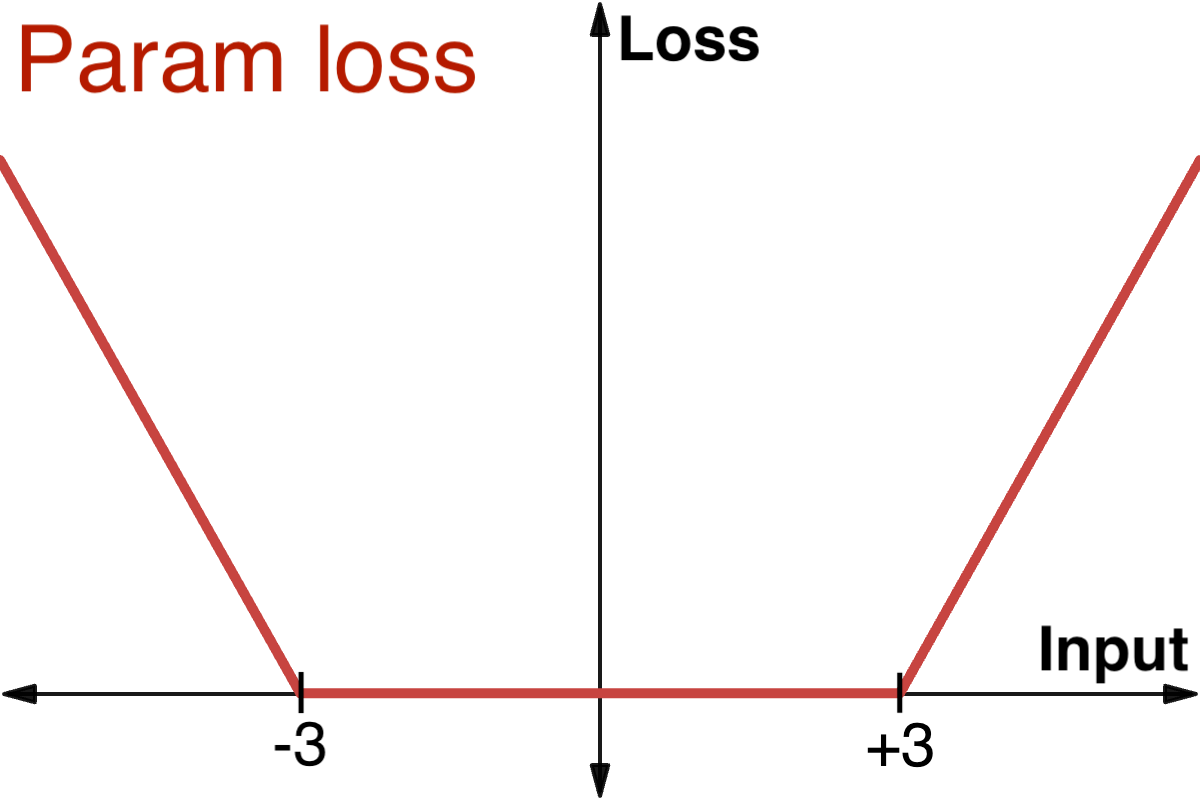}
  \caption{\textbf{Left:} The hard-sigmoid function constrains parameters to $[p_{min}, p_{max}]$, without adding nonlinearity. \textbf{Right:} parameter violation loss pushes the hard-sigmoid input away from flat regions.}
  \label{hardsigmoid_plot}
\end{figure}

Instead, we use a piecewise linear \emph{hard-sigmoid} function (Figure \ref{hardsigmoid_plot} left) to constrain the parameters:
\begin{equation}
p_i = \begin{cases}
            p_i^{min} & \text{if~} \tilde{p}_i \le -3, \\
            p_i^{max} & \text{if~} \tilde{p}_i \ge +3, \\
            \dfrac{p_i^{max} - p_i^{min}}{6} \cdot \tilde{p}_i + \dfrac{p_i^{max} + p_i^{min}}{2} & \text{otherwise}
        \end{cases}
\end{equation}

While the gradient $\frac{\partial p_i}{\partial \tilde{p}_i}$ is zero when $\tilde{p}_i$ is outside the range $[-3, 3]$, we can add ``parameter violation loss'' that places a penalty when $\tilde{p}_i$ is in the flat area (figure \ref{hardsigmoid_plot} right).
\begin{equation}
    \mathcal{L}_{param} = \sum_{i=1}^P \max(0, -\tilde{p}_i - 3, \tilde{p}_i - 3)
\end{equation}
This provides a gradient that pushes $\tilde{p}_i$ towards the linear range $[-3, 3]$ when it is in the flat range. 

For the ecosystem respiration model, we only know that the latent parameter is nonnegative, so we use ReLU to impose the constraint, and a flipped ReLU loss to push inputs out of the flat region.
\begin{equation}
p_i = \max(\tilde{p}_i, 0); \quad \mathcal{L}_{param} = \max(-\tilde{p}_i, 0)
\end{equation}

\subsection{Differentiable Process-Based Decoder}

ScIReN uses a process-based model that expresses output variables as a \emph{fixed, differentiable function} of scientific  parameters and input variables: $\hat{Y} = g_{PBM}(\mathbf{p}, \mathbf{x})$. Two examples are described below.

\paragraph{Ecosystem respiration.} Consider the model of ecosystem respiration in \cite{reichstein2022combining}. Based on scientific knowledge, we write the output variable $R_{eco}$ (ecosystem respiration) as a differentiable function of two latent parameters, base respiration $R_b$ and temperature sensitivity $Q_{10}$:
\begin{equation} \label{reco}
R_{eco} = g_{PBM}(\mathbf{p}, \mathbf{x}) = R_b(\mathbf{x}) \cdot Q_{10}^{\frac{t_a - T_{ref}}{10}}
\end{equation}
where the latent parameters are $\mathbf{p} = \{R_b, Q_{10} \} $, the input features are $\mathbf{x} = \{sw\_pot, dsw\_pot, t_a\}$, and $T_{ref}=15$. $R_b$ has an unknown relationship with the input features (learned in the encoder), while $Q_{10}$ is a learnable constant.

\paragraph{Soil carbon modeling.} As a more complex process, we use the soil organic carbon module from Community Land Model 5 (CLM5) \cite{lu2020full}, which tracks the amount of soil organic carbon (SOC) in 140 pools in the soil (20 depths and 7 material types per depth). Denote the amount of carbon in pool $i$ as $Y_i$. The core of the model is a mass conservation equation for each pool, where the change in carbon equals inflow (from plants and other pools) minus outflow (to other pools or the atmosphere):
\begin{equation}
\frac{d Y_i(t)}{dt} = \text{inflow to pool } i - \text{outflow from pool } i
\end{equation}
The equations for each pool can be combined into a single matrix equation. If we assume steady state ($\frac{d Y_i(t)}{dt} = 0$), we can write $Y$ as a function of biogeochemical parameters $\mathbf{p}$ and input features $\mathbf{x}$:
\begin{equation}
\hat{Y}(\mathbf{p}, \mathbf{x}) = \left[A(\mathbf{p}) \ (\xi(\mathbf{p}, \mathbf{x}) \odot  K(\mathbf{p})) + V(\mathbf{p}, \mathbf{x}) \right]^{-1} I(\mathbf{p}, \mathbf{x})
\end{equation}  
The details of this equation are explained in Appendix \ref{appendix_clm5}. For now, it is sufficient to note that the process-based model takes in 21 latent biogeochemical parameters $\mathbf{p}$, uses the parameters to construct matrices describing  carbon fluxes and decomposition, and finally predicts the amount of carbon in 140 pools (20 layers and 7 pools), $\hat{Y}$. Each operation (including matrix inversion) is differentiable and can be implemented in PyTorch. Note that our labeled data only contains aggregate SOC amounts at specific depths (which may not match the 20 fixed layers). Thus, we sum up the SOC pools at each layer, and linearly interpolate to predict SOC at the observed depths.

\subsection{Final Loss}

The final loss contains a smooth L1 loss between the predicted and ground-truth output variables, as well as the parameter violation and KAN regularization losses:
\begin{align*}
\mathcal{L} = \sum_{i=1}^N &\left[ SmoothL1(\hat{Y_i}, Y_i) \right] + \lambda_{param} \mathcal{L}_{param} + \lambda_{L1} \mathcal{L}_{L1} \\
&+ \lambda_{entropy} \mathcal{L}_{entropy} + \lambda_{smooth} \mathcal{L}_{smooth}
\end{align*}
Since the entire network is differentiable, we can backpropagate the loss through the process-based model to optimize the latent parameters and the learnable weights of the neural network. The loss weights $\lambda$ are tuned on a validation set for each domain; they are intuitive to tune since we can visualize whether the KAN is too sparse/dense and whether the functions are too jagged/smooth. 

\section{Experiments}

We test our approach on two domains: the model of ecosystem respiration in \cite{reichstein2022combining}, and the CLM5 soil carbon cycle model \cite{xu2025biogeochemistry}. 

\subsection{Evaluation Metrics and Baselines}

For each dataset, we first verify that methods are able to predict observed variables, by measuring $R^2$ on the test set. However, the main novelty of ScIReN is its ability to infer hidden functional relationships between input features and latent parameters. We thus conduct experiments on synthetic data to verify that ScIReN infers the correct latent parameters and functional relationships.
To evaluate functional relationship quality, for both ground-truth (synthetic) relationships and our learned models, we first compute the fraction of variance in the output that is explained by each input feature. For 1-layer KAN, we can simply compute the variance of each edge's post-activation outputs, and divide by the total variance in the output. For other models, this is non-trivial; we use Partial Dependence Variance \cite{greenwell2018simple} to estimate this. We then use KL divergence to measure how far the model's learned feature importance distributions are from the ground-truth. 

 For baselines, we compare against a pure neural network that only predicts observed variables (and cannot infer latent variables), and a blackbox-hybrid model \cite{reichstein2022combining,xu2025biogeochemistry} where a neural network (or linear model) predicts latent parameters, which are passed through the process-based model. We run ScIReN and Blackbox-Hybrid using a nonlinear constraint (sigmoid or softplus) and our proposed linear constraint (hard-sigmoid or ReLU).

\subsection{Ecosystem Respiration}

\begin{table*}
  \centering
  \begin{tabular}{lccc}
    \toprule
    \textbf{Method} & $R^2$ (observed, $\uparrow$) & $R^2$ (latent, $\uparrow$) & KL, functional relationships $(\downarrow)$ \\ \hline
    Pure-NN & 0.968 $\pm$ 0.004 & N/A & N/A \\ 
    Blackbox-Hybrid, nonlinear constraint & 0.972 $\pm$ 0.003 & 0.818 $\pm$ 0.052 & 0.284 $\pm$ 0.041 \\
    Blackbox-Hybrid, linear constraint & \textbf{0.976} $\pm$ 0.001 & 0.941 $\pm$ 0.031 & 0.161 $\pm$ 0.068 \\
    ScIReN, nonlinear constraint (1-layer KAN) & 0.975 $\pm$ 0.000 & 0.995 $\pm$ 0.000 & 0.002 $\pm$ 0.001 \\
    ScIReN, linear constraint (1-layer KAN) & \textbf{0.976} $\pm$ 0.000 & \textbf{1.000} $\pm$ 0.000 & \textbf{0.001} $\pm$ 0.001 \\ \hline
  \end{tabular}
  \caption{Ecosystem respiration, linear $R_b$. Mean and standard deviation across 5 seeds.}
  \label{result1}
\end{table*}

\begin{table*}
  \centering
  \begin{tabular}{lccc}
    \toprule
    \textbf{Method} & $R^2$ (observed, $\uparrow$) & $R^2$ (latent, $\uparrow$) & KL, functional relationships $(\downarrow)$ \\ \hline
    Pure-NN & 0.946 $\pm$ 0.004 & N/A & N/A \\ 
    Blackbox-Hybrid, nonlinear constraint & 0.948 $\pm$ 0.008 & -0.265 $\pm$ 0.355 & 0.937 $\pm$ 0.168 \\
    Blackbox-Hybrid, linear constraint & \textbf{0.961} $\pm$ 0.001 & 0.949 $\pm$ 0.059 & 0.199 $\pm$ 0.094 \\
    Linear-Hybrid, linear constraint & 0.400 $\pm$ 0.002 & -0.204 $\pm$ 0.080 & 1.078 $\pm$ 0.115 \\
    ScIReN, linear constraint (1-layer KAN) & 0.651 $\pm$ 0.014 & -1.225 $\pm$ 0.668 & 1.182 $\pm$ 0.085 \\
    ScIReN, linear constraint (2-layer KAN) & \textbf{0.961} $\pm$ 0.001  & \textbf{0.993} $\pm$ 0.005 & \textbf{0.078} $\pm$ 0.045 \\ \hline
  \end{tabular}
  \caption{Ecosystem respiration, nonlinear $R_b$. Mean and standard deviation across 5 seeds.}
  \label{result2}
\end{table*}

For ecosystem respiration, we used the same dataset and splits as \cite{reichstein2022combining}, except we removed the 20\% highest-temperature examples from the train set, forcing the model to extrapolate to higher temperatures than seen during training. We created two sets of latent $R_b$ (base respiration) values. First, we model $R_b$ as a linear function of 2 features $sw\_pot$, $dsw\_pot$ (as in \citet{reichstein2022combining}):
    \begin{equation*} R_b = 0.0075 \cdot sw\_pot -  0.00375 \cdot dsw\_pot + 1.03506858 \end{equation*}
Second, to create a setting where a 2-layer KAN is needed, we add an absolute value.
    \begin{equation}R_b' = 0.0075 \cdot sw\_pot -  0.00375 \cdot dsw\_pot
    \end{equation}
    \begin{equation} R_b = \left| \frac{R_b' - mean(R_b')}{stdev(R_b')} \right| + 0.1
    \end{equation}
We then generate the observed variable $R_{eco}$ using the process-based model (Eq. \ref{reco}) with multiplicative noise:
\begin{equation}
    R_{eco} = R_b \cdot Q_{10}^{\frac{t_a - T_{ref}}{10}} \cdot (1+\epsilon), \quad \epsilon \sim N(0, 0.1),
\end{equation}  
Table \ref{result1} shows results for the first setting (linear $R_b$). For predicting the observed variable $R_{eco}$, Blackbox-Hybrid and ScIReN outperform Pure-NN as the process-based model provides prior knowledge that helps the model extrapolate out-of-distribution. For inferring the latent variable $R_b$ and functional relationships, ScIReN with linear constraint does best; it correctly learns that $R_b$ only depends on $sw\_pot$ and $dsw\_pot$, not $t_a$ (see Figure \ref{encoder} left). This is difficult because the irrelevant feature $t_a$ (air temperature) is highly correlated with feature $sw\_pot$. Also, if the model learns the wrong $Q_{10}$ value, it can make $R_b$ depend on $t_a$ to compensate. ScIReN's entropy loss pushes it to eliminate as many variables as possible, and the smoothness loss (with the linear constraint) makes the relationship as linear as possible. Other methods learn complex relationships that perform worse.

Table \ref{result2} shows results for nonlinear $R_b$, where a 2-layer KAN is  needed to model the complex relationship (1 layer is insufficient). ScIReN predicts the observed variable, latent variable, and functional relationships almost perfectly, greatly outperforming Pure-NN and Blackbox-Hybrid. Figure \ref{encoder} (right) shows that ScIReN learned the true relationship.


\begin{table*}
  \centering
  \begin{tabular}{lccc}
    \toprule
    \textbf{Method} & $R^2$ (observed, $\uparrow$) & $R^2$ (latent, $\uparrow$) & KL, functional relationships $(\downarrow)$ \\ \hline
    Pure-NN & 0.933 $\pm$ 0.015 & N/A & N/A \\
    Blackbox-Hybrid, nonlinear constraint & 0.996 $\pm$ 0.003 & 0.226 $\pm$ 0.800 & 1.312 $\pm$ 0.170 \\
    Blackbox-Hybrid, linear constraint & 0.995 $\pm$ 0.003 & 0.721 $\pm$ 0.226 & 1.082 $\pm$ 0.258 \\
    Linear-Hybrid, linear constraint & 0.979 $\pm$ 0.013 & 0.087 $\pm$ 1.014 & 1.727 $\pm$ 0.322 \\
    ScIReN, linear constraint (1-layer KAN) & \textbf{0.999} $\pm$ 0.002 & \textbf{0.989} $\pm$ 0.019 & \textbf{0.080} $\pm$ 0.042 \\ \hline
  \end{tabular}
  \caption{Soil carbon cycle (synthetic parameters). Mean and standard deviation across 5 splits/seeds.}
  \label{result3}
\end{table*}



\begin{figure*}
  \centering
  \includegraphics[height=0.35\textwidth,trim={0 0 0 -0.5cm, clip}]{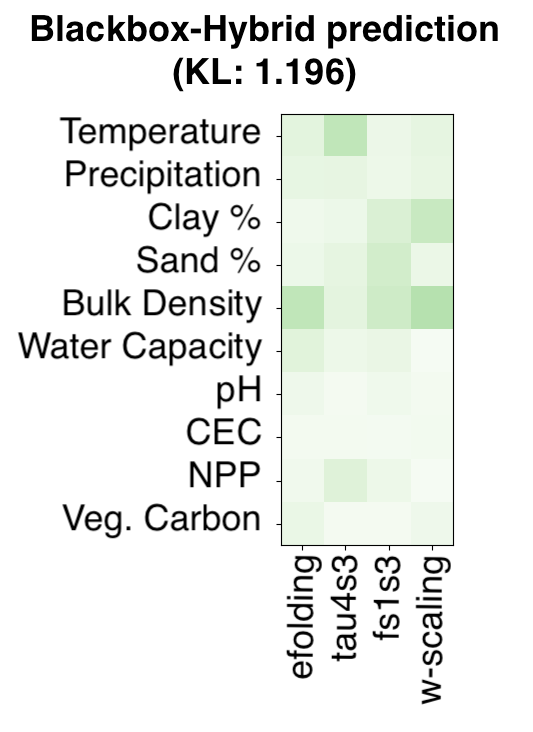}
  \includegraphics[height=0.35\textwidth,trim={0 0 0 0cm, clip}]{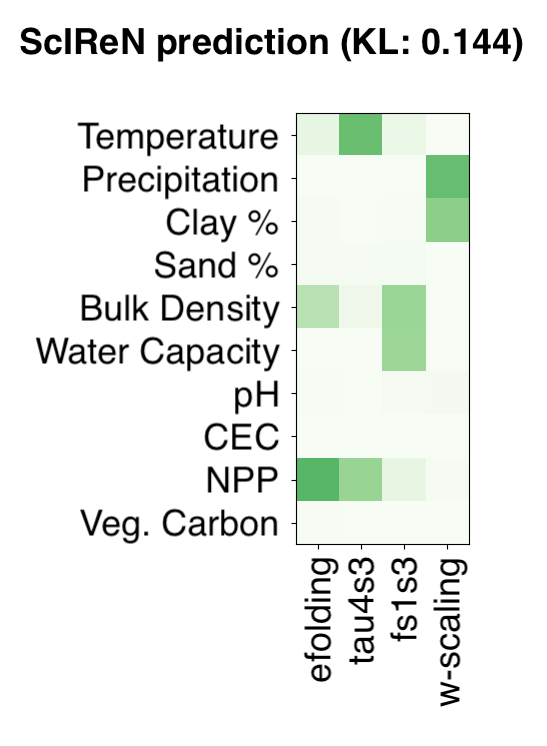}
  \includegraphics[height=0.35\textwidth,trim={0 0 0 0cm, clip}]{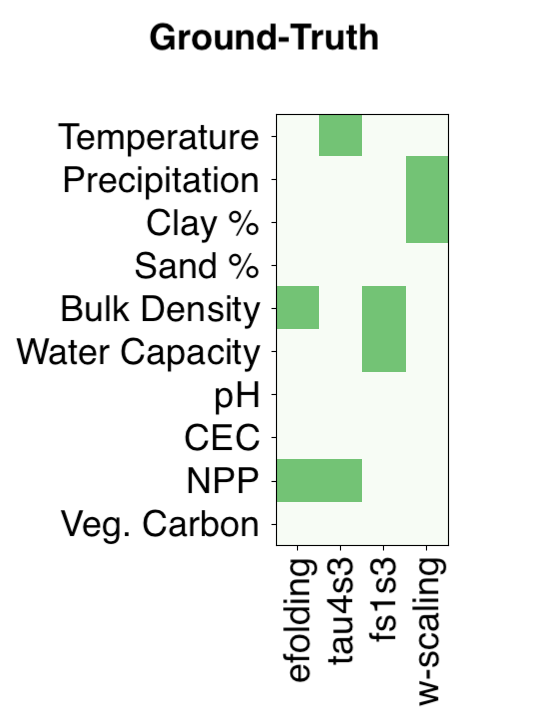}
  \caption{Functional relationships learned by Blackbox-Hybrid (left) and ScIReN (center) vs. truth (right), on synthetic labels. ScIReN recovers the true relationships much more accurately.}
  \label{functional}
\end{figure*}

\begin{table*}
  \centering
  \begin{tabular}{lccc}
    \toprule
    \textbf{Method} & $R^2 (\uparrow)$ & MAE $(\downarrow)$ & Pearson correlation $(\uparrow)$\\ \hline
    Pure-NN & 0.552 $\pm$ 0.173 & \textbf{4609.3} $\pm$ 356.8 & \textbf{0.780} $\pm$ 0.053 \\
    Blackbox-Hybrid, nonlinear constraint & 0.584 $\pm$ 0.082 & 4726.2 $\pm$ 727.3 & 0.776 $\pm$ 0.048 \\
    Blackbox-Hybrid, linear constraint & \textbf{0.589 $\pm$ 0.070} & 4849.7 $\pm$ 650.3 & 0.774 $\pm$ 0.040
    \\
    Linear-Hybrid, linear constraint & 0.552 $\pm$ 0.082 & 4984.8 $\pm$ 771.6 & 0.761 $\pm$ 0.046 \\
    ScIReN, linear constraint (1-layer KAN) & 0.582 $\pm$ 0.080 & 4708.2 $\pm$ 673.1 & 0.769 $\pm$ 0.049 \\
    ScIReN, linear constraint (2-layer KAN) & 0.571 $\pm$ 0.094 & 4707.3 $\pm$ 826.3 & 0.765 $\pm$ 0.052 \\ \hline
  \end{tabular}
  \caption{Soil carbon cycle (real labels). Mean and standard deviation across 5 splits/seeds.}
  \label{result4}
\end{table*}

\subsection{Soil Carbon Cycle}

For soil carbon, we split the US into $2 \times 2$ degree blocks and randomly assign the blocks to five folds, as in \cite{wang2020weakly}.
We average across five data splits -- each split uses one fold for testing, one for validation, and the other folds for training. Each split also uses its own initial seed. 

First, we synthetically generate functional relationships between the 10 input features and 4 most sensitive biogeochemical parameters from \cite{xu2025biogeochemistry}. Of the $10\times 4 = 40$ possible relationships, we select $20\%$ and randomly assign each to (linear, quadratic, log, exp, abs) with random affine shifts. We set the other parameters to default values as they are poorly constrained by data \cite{xu2025biogeochemistry}; this \emph{equifinality} is an inherent limitation of process-based models, which we mitigate by only predicting the 4 most sensitive parameters. We then use the CLM5 process-based model to generate synthetic SOC labels from these parameters. Table \ref{result3} shows how various methods perform in recovering these functional relationships. ScIReN (1-layer KAN) recovers the ground-truth relationships (see Figure \ref{functional}) and observed/latent variables almost perfectly, while Blackbox-Hybrid is not incentivized to produce sparse relationships and mixes correlated features in.

Finally, we train all methods on real carbon labels in Table \ref{result4}. ScIReN's accuracy in predicting SOC amounts is comparable to Blackbox-Hybrid and outperforms pure neural networks (in terms of $R^2$). This indicates that we can obtain full interpretability without significantly sacrificing predictive accuracy. Note that a 1-layer KAN is enough to achieve good accuracy, making the encoder even easier to interpret. Qualitatively, the predicted relationships are consistent with ecological knowledge; see Appendix \ref{qualitative} for details.




\section{Conclusion}

We have proposed ScIReN, an end-to-end framework that combines data-driven learning with established scientific knowledge (in the form of process-based reasoning) to discover interpretable relationships between latent scientific parameters and inputs. The entropy loss helps the model discover robust relationships and disregard irrelevant variables, while the smoothness loss enables expressivity without overfitting. A limitation of ScIReN is that the KAN encoder is somewhat sensitive to initialization and hyperparameters; future work could pursue making KANs easier to train. 
Nevertheless, ScIReN demonstrates excellent performance at retrieving true functional relationships, can extrapolate out-of-distribution, and contains useful biases towards simplicity. Most importantly, ScIReN is fully transparent -- it respects existing scientific knowledge  while generating novel insights, thus facilitating data-driven scientific discovery.


\section*{Acknowledgements}

This research is supported by AI-LEAF: “AI Institute for Land, Economy, Agriculture, and Forestry”, funded by the USDA National Institute of Food and Agriculture (NIFA) and the NSF National AI Research Institutes Competitive Award. This research is also partially supported by Schmidt Sciences programs, an AI2050 Senior Fellowship and an Eric and Wendy Schmidt AI in Science Postdoctoral Fellowship; the National Science Foundation; the US Department of Energy; the CALS Moonshot Seed Grant program; the “NYS Connects: Climate Smart Farms \& Forestry” project, funded by the USDA, the New York State Department of Environmental Conservation, and the New York State Department of Agriculture and Markets; the National Science Foundation (NSF) through an NSF Research Traineeship (NRT) project in AI for Sustainability under Grant No. 2345579;  and the Air Force Office of Scientific Research.


\medskip

\bibliography{ref}

\newpage



\clearpage

\makeatletter
\@ifundefined{isChecklistMainFile}{
  \newif\ifreproStandalone
  \reproStandalonetrue
}{
  \newif\ifreproStandalone
  \reproStandalonefalse
}
\makeatother

\ifreproStandalone
\documentclass[letterpaper]{article}
\usepackage[submission]{aaai2026}
\setlength{\pdfpagewidth}{8.5in}
\setlength{\pdfpageheight}{11in}
\usepackage{times}
\usepackage{helvet}
\usepackage{courier}
\usepackage{xcolor}
\usepackage{amsmath}
\usepackage{amsfonts}
\usepackage{natbib}
\usepackage{booktabs}
\usepackage{graphicx}
\frenchspacing

\begin{document}
\fi

\appendix

\section{KAN edge score computation} \label{edge_score}

We review how we compute edge scores in KAN for sparsity regularization, which is similar to the KAN 2.0 paper \cite{liu2024kan2}.

Define $E_{l,i,j}$ as the mean absolute deviation\footnote{Mean absolute deviation is similar to standard deviation, but is the average L1 distance to the mean, instead of the square root of the average squared distance.})
of the outputs of the $(l,i\rightarrow j)$ edge (the edge from layer $l-1$, node $i$ to layer $l$, node $j$):
\begin{equation} 
E_{l,i,j} = \text{AbsDev}(\phi_{l,i,j}(x_{l-1, i}))
\end{equation}
Note that the mean absolute deviation is taken over the \textbf{batch} dimension.

Let $N_{l, j}$ be the mean absolute deviation of the outputs of node $(l, j)$:
\begin{equation}
    N_{l,j} = \text{AbsDev} \left( \sum_{i=1}^{n_{l-1}} \phi_{l,i,j}(x_{l-1, i}) \right)
\end{equation} 
We now compute node and edge scores iteratively. Start with last layer, and set output node scores  $A_{L,i}$ to be the variance of output $i$. Then compute scores as follows for each layer $l = L, \dots, 1$:\footnote{Eq 9a in KAN 2.0 paper wrote the first equation differently, but I think it must be typos? This is my guess based on intuition and code.}
\begin{equation}
    B_{l-1, i, j} = A_{l, j} \frac{E_{l-1,i,j}}{N_{l, j}}
\end{equation}
\begin{equation}
    A_{l-1, i} = \sum_{j=0}^{n_l} B_{l-1, i, j}
\end{equation}

Intuitively, $A_{l,j}$ represents how much neuron $(l,j)$ contributes to the variance in all final outputs, and $B_{l,i,j}$ is how much of that variance is contributed by the output of edge $(l, i\rightarrow j)$. For the first equation, we first look at the contribution of neuron $(l,j)$ towards the final variances, and then split it across the input edges according to the fraction of this neuron's variance contributed by each incoming edge ($\frac{E_{l-1,i,j}}{N_{l,j}}$). For the second equation, we compute each neuron's contribution towards the final variances by summing over the contributions via each \emph{outgoing} edge.

\section{Ecosystem respiration: data and process-based model}

For the ecosystem respiration experiments, we used the process-based model and dataset from  \cite{reichstein2022combining}. We have labels for output variable $R_{eco}$ (ecosystem respiration). The process-based model specifies that $R_{eco}$ is a differentiable function of two latent parameters, base respiration $R_b$ and temperature sensitivity $Q_{10}$. Specifically:
\begin{equation}
R_{eco} = g_{PBM}(\mathbf{p}, \mathbf{x}) = R_b(\mathbf{x}) \cdot Q_{10}^{\frac{t_a - T_{ref}}{10}}
\end{equation}
where the latent parameters are $\mathbf{p} = \{R_b, Q_{10} \} $, the input features are $\mathbf{x} = \{sw\_pot, dsw\_pot, t_a\}$ (potential shortwave radiation, derivative of potential shortwave radiation, air temperature), and $T_{ref}=15$ is a constant. In this example, $R_b$ is a function of input features, while $Q_{10}$ is assumed to be independent of the input features and is a global learnable constant.

Note that this model is somewhat ill-posed, as the input feature temperature ($t_a$) can influence $R_{eco}$ either through changing the base respiration $R_b$, or through the $Q_{10}$ exponential term. For example, as explained in \cite{cohrs2024causal}, if the model predicted the wrong value of $Q_{10}$, it can adjust its predicted $R_b$ to compensate and still match the labels for $R_{eco}$, as follows:
\begin{equation}
    R_{eco} = \underbrace{ R_b^{TRUE}(\mathbf{x}) \cdot \left(\frac{Q_{10}^{TRUE}}{Q_{10}^{PRED}} \right)^{\frac{t_a - T_{ref}}{10}} }_{\text{Predicted } R_b}\cdot \left( Q_{10}^{PRED} \right)^{\frac{t_a - T_{ref}}{10}}
\end{equation}

where the predicted $R_b$ function soaked up additional temperature dependencies that should have been included in the $Q_{10}$ exponential. Thus the task is difficult, but ScIReN's bias towards linear and sparse functions can allow it to recover the true $R_b$.

The dataset and splits are the same as used in \cite{reichstein2022combining, cohrs2024causal}, except that we remove datapoints with top-20\% $t_a$ (temperature) values from the training set (and not the validation/test sets), to test the model's ability to generalize out-of-distribution. This results in 56102 training, 17520 validation, and 17568 test datapoints. The input features are taken from values observed by a EC tower in Neustift, Austria.  The $Q_{10}$ parameter (latent) was set to 1.5. We created two sets of latent $R_b$ (base respiration) values:
\begin{enumerate} 
    \item First, we model $R_b$ as a linear function of 2 features $sw\_pot$, $dsw\_pot$ (following \cite{reichstein2022combining}):
    \begin{equation} 
    R_b = 0.0075 \cdot sw\_pot -  0.00375 \cdot dsw\_pot + 1.03506858
    \end{equation}
    where the constant $1.03506858$ was selected to make all values non-negative.
    \item Second, to create a setting where 2-layer KAN is needed, we add an absolute value.
    \begin{align}
    R_b' &= 0.0075 \cdot sw\_pot -  0.00375 \cdot dsw\_pot \\
    \quad R_b &= \left| \frac{R_b' - mean(R_b')}{stdev(R_b')} \right| + 0.1
    \end{align}
\end{enumerate}
We then generated the observed variable $R_{eco}$ according to the process-based model with multiplicative noise, as in \cite{cohrs2024causal}:
\begin{align}
    &R_{eco} = R_b \cdot Q_{10}^{\frac{t_a - T_{ref}}{10}} \cdot (1+\epsilon), \nonumber \\
    &\epsilon \sim N(0, 0.1), \text{truncated to } [-0.95, 0.95]
\end{align}

\section{Ecosystem respiration: hyperparameters} \label{hyp_eco}

We initialized $Q_{10}$ to 0.5 for all models (so it should increase during training), and give $Q_{10}$ a learning rate that is 100 times as high as the other learnable parameters in the model. We fixed $\lambda_{param}$ to 1 if using the ReLU constraint. For Pure-NN and Blackbox-Hybrid, we used an MLP with 2 layers and 16 hidden units. For ScIReN, we used KANs with cubic splines, 30 grid points, and a margin of 2 times the input range.

For each method, we tuned hyperparameters using a grid search, selecting the hyperparameters that had the lowest validation mean-squared-error loss. For Pure-NN and Blackbox-Hybrid, we considered learning rates from $[10^{-4}, 10^{-3}, 10^{-2}, 10^{-1}]$, and weight decay from $[0, 10^{-4}, 10^{-3}]$. For the KAN-based methods, we considered learning rates from $[10^{-3}, 10^{-2}, 10^{-1}]$, weight decay from $[0, 10^{-4}]$, and considered $\lambda_{entropy}, \lambda_{L1}, \lambda_{smooth}$ from $[10^{-3}, 10^{-2}, 10^{-1}, 1]$. (We generally set $\lambda_{entropy}=\lambda_{L1}$. Due to the number of hyperparameters in KAN, we did not try all configurations.)

The hyperparameters selected for Table 1 (linear $R_b$) are shown below in Table \ref{hyperparam1}.  We also report approximate runtimes per run on a single V100 GPU.

\begin{table*}
  \caption{Hyperparameters for Table 1.}
  \label{hyperparam1}
  \centering
  \begin{tabular}{lcccc|c}
    \toprule
    \textbf{Method} & LR & Weight decay & $\lambda_{entropy}, \lambda_{L1}$ & $\lambda_{smooth}$ & Runtime \\ \hline
    Pure-NN & 0.1 & 0 & - & - &  15min \\ 
    Blackbox-Hybrid, nonlinear constraint & 0.1 & 0 & - & - & 24min \\
    Blackbox-Hybrid, linear constraint & 0.1 & 0 & - & - & 24min \\
    ScIReN, nonlinear constraint (1-layer KAN) & 0.1 & $10^{-4}$ & 0.01 & 0.1 & 15min \\
    ScIReN, linear constraint (1-layer KAN) &  0.01 & $10^{-4}$ & 0.01 & 1.0 & 15min \\ \hline
  \end{tabular}
\end{table*}

The hyperparameters selected for Table 2 (nonlinear $R_b$) are shown below in Table \ref{hyperparam2}, along with approximate runtimes per run on a single V100 GPU. Note that for 2-layer KAN, we used 8 hidden units, and we used a `zero` base function (instead of identity).

\begin{table*}
  \caption{Hyperparameters for Table 2.}
  \label{hyperparam2}
  \centering
  \begin{tabular}{lcccc|c}
    \toprule
    \textbf{Method} & LR & Weight decay & $\lambda_{entropy}, \lambda_{L1}$ & $\lambda_{smooth}$ & Runtime \\ \hline
    Pure-NN & 0.1 & 0 & - & - &  25min \\ 
    Blackbox-Hybrid, nonlinear constraint & $10^{-3}$ & $10^{-3}$ & - & - & 25min \\
    Blackbox-Hybrid, linear constraint & $10^{-3}$ & $10^{-4}$ & - & - & 25min \\
    ScIReN, linear constraint (1-layer KAN) & 0.1 & $10^{-4}$ & 0.01 & 0.1 & 20min \\
    ScIReN, linear constraint (2-layer KAN) &  0.01 & $10^{-4}$ &  $10^{-3}$ & 1.0 & 38min \\ \hline
  \end{tabular}
\end{table*}

\section{Ecosystem respiration: Additional figures}

\begin{figure*}
  \centering
  \includegraphics[width=0.75\textwidth]{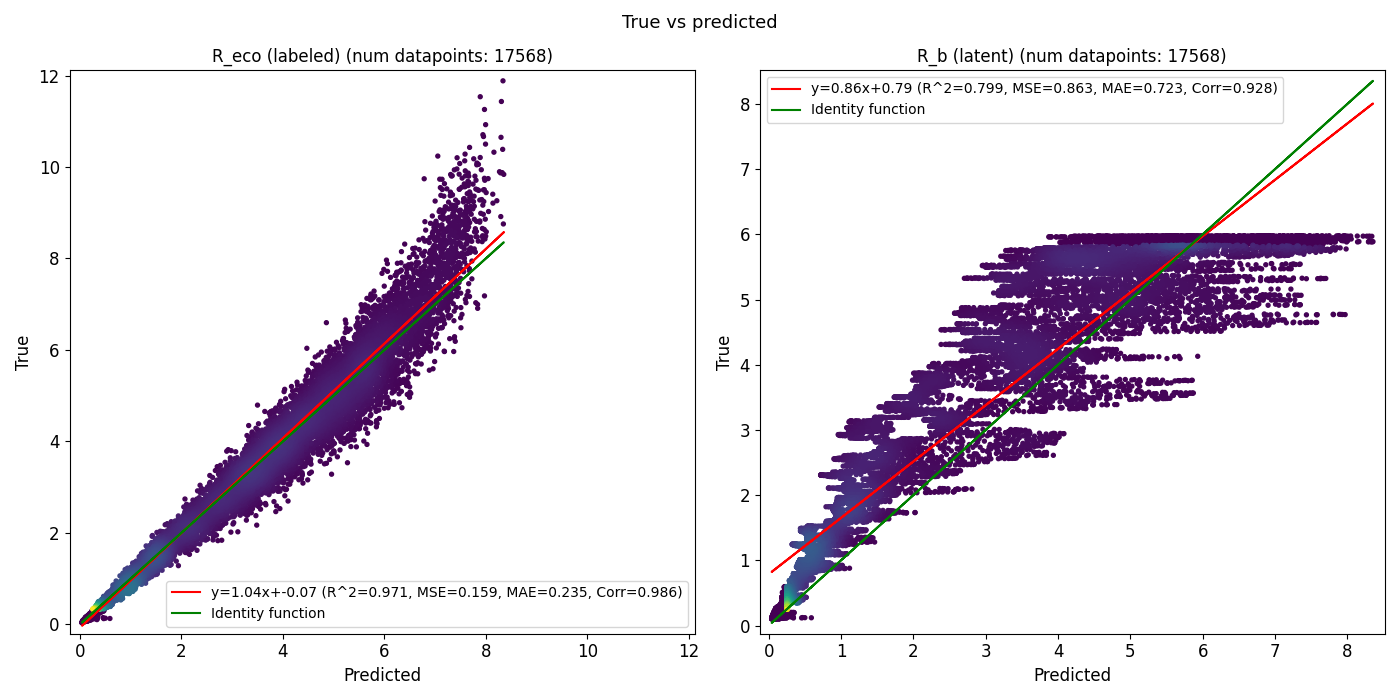}
  \caption{Pure NN (Table 1, linear $R_b$). Because the model does not have access to scientific knowledge (process-based model), it struggles to extrapolate on the higher end of ecosystem respiration / temperature (left). It is also unable to predict the latent variable (right).}
  \label{plot_1a}
\end{figure*}

\begin{figure*}
  \centering
  \includegraphics[width=0.75\textwidth]{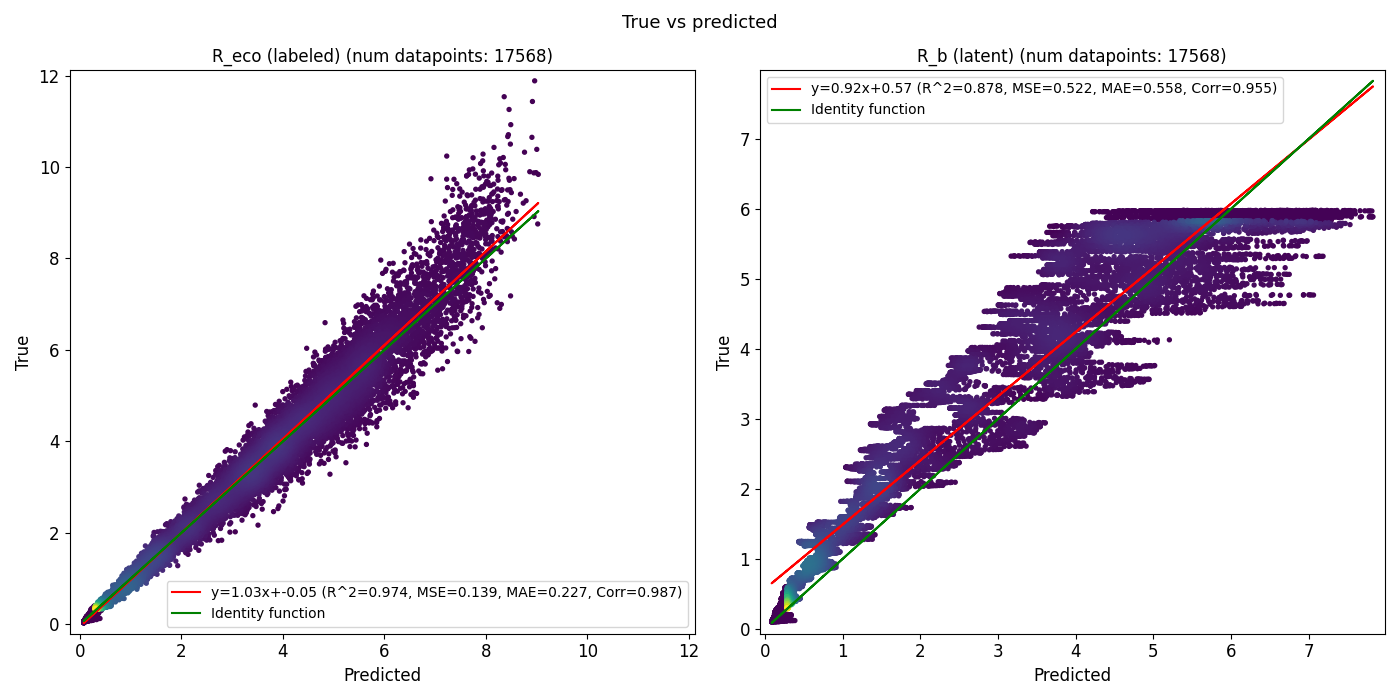}
  \includegraphics[width=0.35\textwidth]{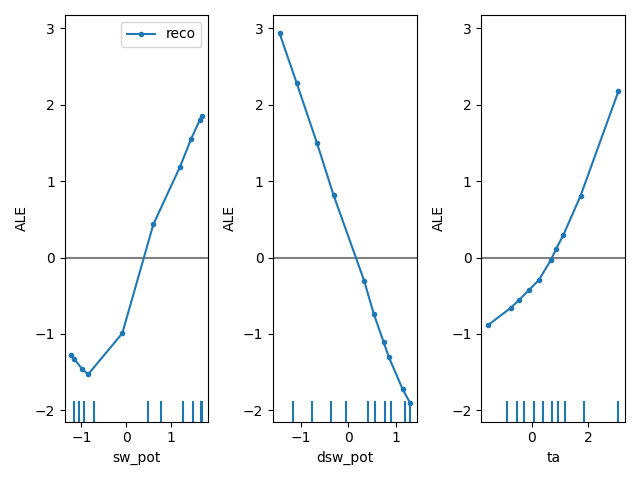}
  \caption{Blackbox-Hybrid, nonlinear constraint (Table 1, linear $R_b$). The process-based model helps with extrapolation (top left). However the latent variable prediction is inaccurate (top right). According to the Accumulated Local Effects plots (which estimate the impact of each input feature; bottom), the model thinks that both $sw\_pot$ and $t_a$ influence the latent variable, but only $sw\_pot$ is actually relevant.}
  \label{plot_1b}
\end{figure*}

\begin{figure*}
  \centering
  \includegraphics[width=0.75\textwidth]{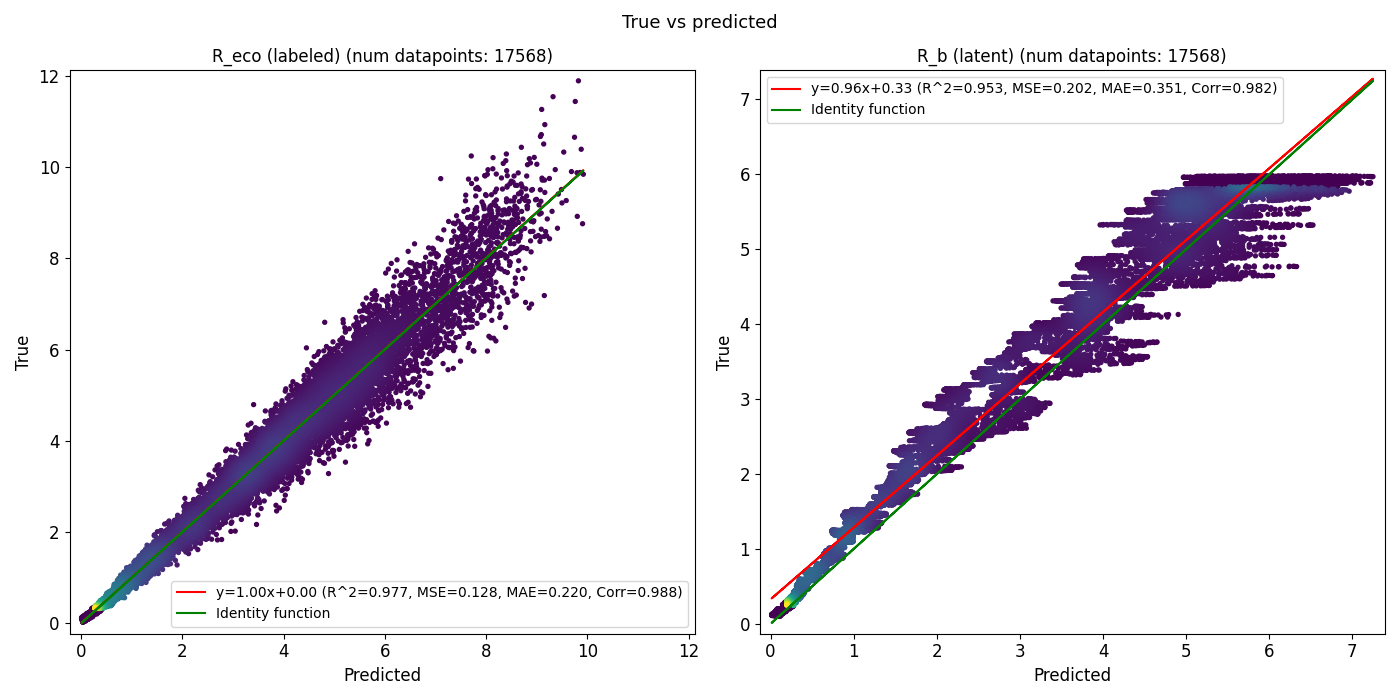}
  \includegraphics[width=0.35\textwidth]{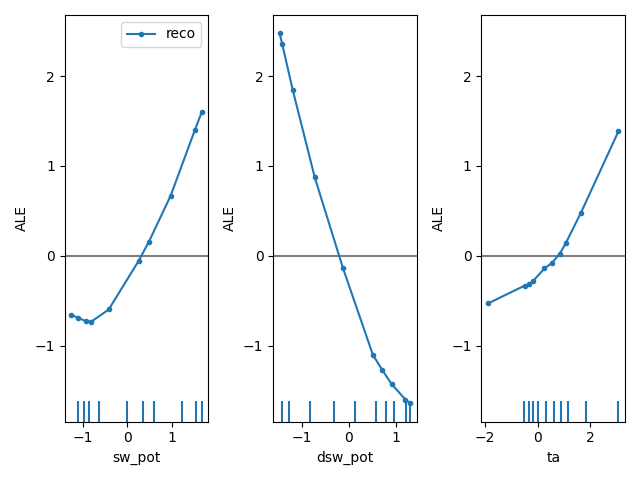}
  \caption{Blackbox-Hybrid, linear constraint (Table 1, linear $R_b$). Changing to linear constraint improves the latent variable prediction (top right) since the model now only needs to learn a simple linear relationship between $R_b$ and input features. However, the model is still using $t_a$ somewhat even though it is actually useless (bottom).}
  \label{plot_1c}
\end{figure*}

\begin{figure*}
  \centering
  \includegraphics[width=0.75\textwidth]{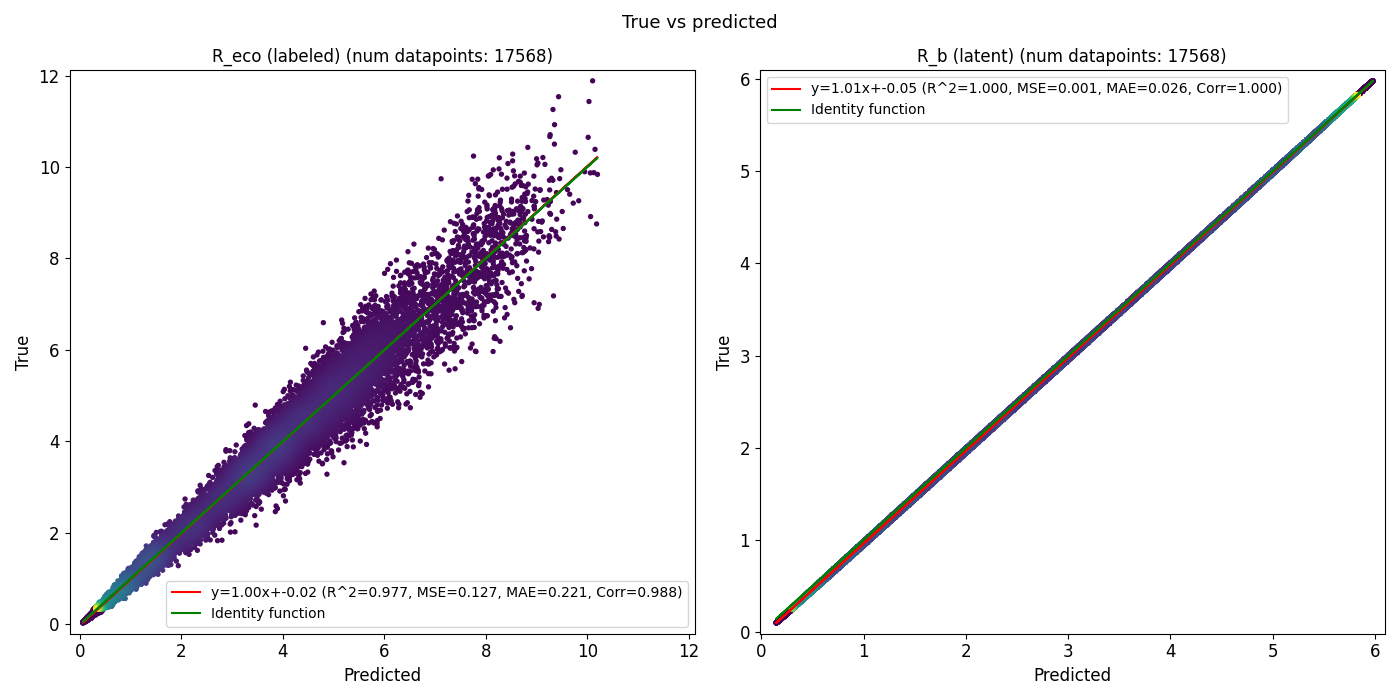}
  \includegraphics[width=0.5\textwidth]{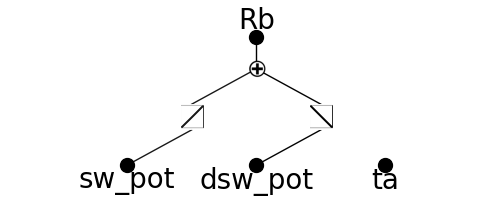}
  \caption{ScIReN, linear constraint (Table 1, linear $R_b$). ScIReN's entropy loss encourages sparsity, forcing it to choose the best among correlated features. The smoothness loss also encourages a near-linear function. Thus, it is able to predict the observed variable (top left), latent variable (top right), and functional relationships (bottom) correctly.}
  \label{plot_1e}
\end{figure*}

For Linear $R_b$ we present additional figures demonstrating the qualitative performance of various methods (Figures \ref{plot_1a}, \ref{plot_1b}, \ref{plot_1c}, \ref{plot_1e}).

\section{Soil carbon cycle: CLM5 process-based model and datasets} \label{appendix_clm5}

The textbook by \citet{luo2022land} provides a detailed introduction to land carbon cycle models, but we summarize some key details here.  As stated in the main text, we can write the change in carbon for each pool as the difference between inputs and outputs:
\begin{equation}
\frac{d Y_i(t)}{dt} = \text{inflow to pool } i - \text{outflow from pool } i
\end{equation}
Since there are 140 equations (one for each pool), we can stack them into a single matrix equation as follows:
\begin{align} 
\frac{dY(t)}{dt} &= B(\mathbf{p}, t) I(t) - A(\mathbf{p}) \ \text{diag}(\xi(t) \odot  K(\mathbf{p})) \  Y(t) \nonumber \\
& \quad - V(\mathbf{p}, t)Y(t)
\end{align}
Here, $Y(t) \in \mathbb{R}^{140}$ is the amount of carbon in each of the 140 pools at time $t$. $I(t) \in \mathbb{R}$ is the total carbon input to the system from vegetation at time $t$, and $B(t) \in \mathbb{R}^{140}$ is the allocation of input to different pools. $A(\mathbf{p}, t) \in \mathbb{R}^{140 \times 140}$ quantifies carbon transfers between pools at the same depth. $K(\mathbf{p}) \in \mathbb{R}^{140}$ contains the intrinsic decomposition rate of each carbon pool, which is the same for each pool across 20 layers. $\xi(t) \in \mathbb{R}^{140}$ is computed based on monthly climate data and indicates how the environment modifies the intrinsic decomposition rate. $\xi$ and $K$ are multiplied elementwise and then converted into a $140\times140$ diagonal matrix. $V(\mathbf{p}, t) \in \mathbb{R}^{140 \times 140}$ quantifies vertical carbon transfers between adjacent depths. $\odot$ denotes elementwise multiplication, while $\text{diag}$ indicates we construct a diagonal matrix from the given vector. The $t$ indicates that the parameter varies with time, while $\mathbf{p}$ indicates that the matrix is constructed in a deterministic way from latent biogeochemical parameters predicted by the encoder. In this study we do not consider temporal variations, but any parameters that depend on $t$ are constructed using \emph{monthly-average} climate forcing data.

In this work, we assume steady state ($\frac{dY}{dt} = 0$), which is a reasonable assumption since previous research has shown that recent disequilibrium effects from climate change and human activities are minor in comparison to the total SOC storage that has developed over thousands of years \cite{tao2023microbial,lu2018ecosystem}. Thus, we can solve for $Y$ analytically as a function of $\mathbf{p}$:
$$\hat{Y}(t) = \left[A(\mathbf{p}) \ \text{diag}(\xi(t) \odot  K(\mathbf{p}))   + V(\mathbf{p}, t) \right]^{-1} B(\mathbf{p}, t) I(t)$$
Note that $\hat{Y}(t)$ contains predicted carbon amounts at 20 fixed layers, and 7 pools per layer. However, our labels are only aggregate SOC amounts at specific depths (which may not match the 20 fixed layers).  Thus, we sum up the SOC pools at each layer, and linearly interpolate to predict SOC at the observed depths.

The datasets used in these experiments are documented in \cite{tao2020deep, xu2025biogeochemistry}. We restricted to ten input features for interpretability:
\begin{enumerate}
    \item Annual mean temperature
    \item Annual precipitation
    \item Clay content
    \item Sand content
    \item Bulk density
    \item Soil water capacity
    \item Soil pH in H2O
    \item Cation Exchange Capacity
    \item NPP (Net Primary Productivity)
    \item Vegetation Carbon Stock
\end{enumerate}

To test ScIReN's ability on small datasets, we only use 1018 examples in our dataset total, which are representative locations in the conterminous United States. We split the dataset spatially into $2\times2$-degree blocks, and assign each block to one of 5 folds. We run experiments across 5 splits, each time holding out a different test fold (and validation fold). 

For the synthetic functional relationship experiment, we only considered the four most sensitive biogeochemical parameters identified in \cite{xu2025biogeochemistry}, as the other parameters are poorly constrained by data. We randomly selected 20\% of the possible input-output relationships to exist. For each relationship, we randomly draw a function (linear, quadratic, exponential, logarithmic, absolute value), and add random affine shifts. We then calculate the value of each output parameter as the sum of functions of each relevant input, e.g.
\begin{equation}
p_i = f_{i,2}(x_2) + f_{i,6}(x_6) + \epsilon =\log(ax_2 + b) + (cx_6^2 + d)
\end{equation}

\begin{figure*}
  \centering
  \includegraphics[width=0.9\textwidth]{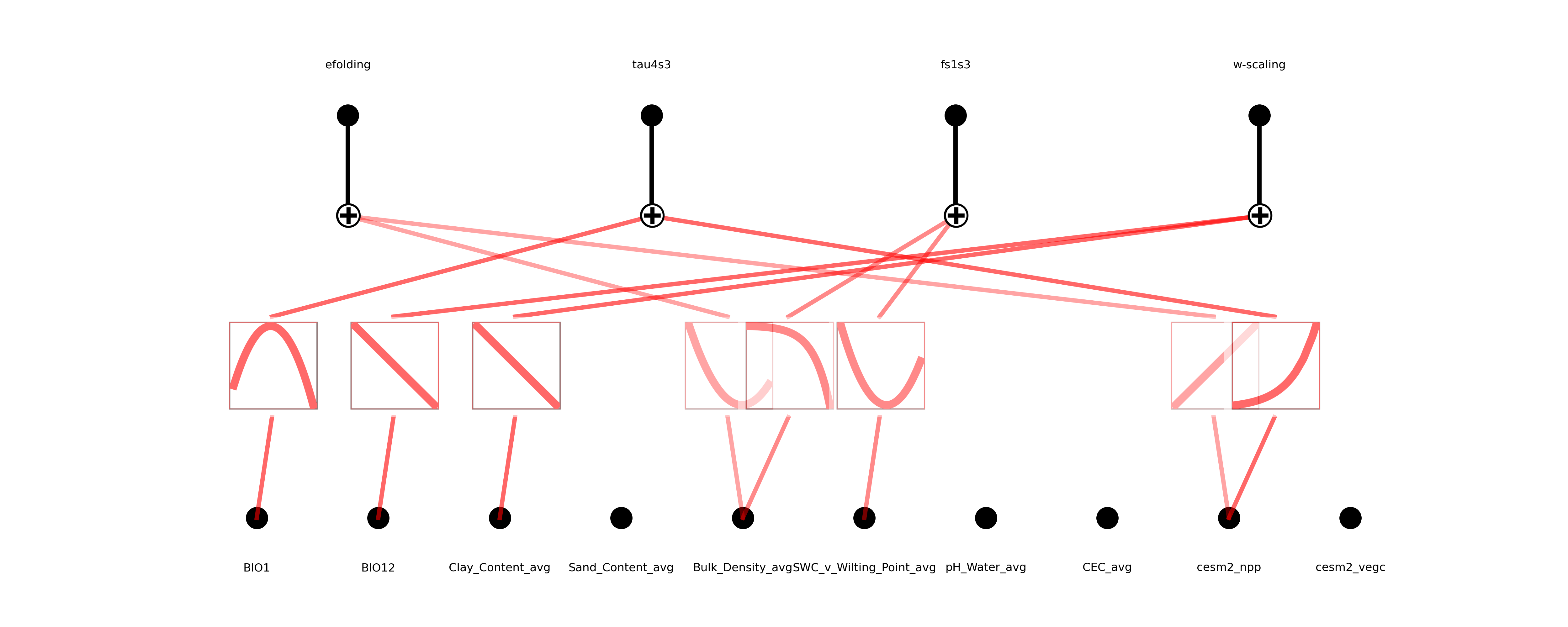}
  \caption{Prescribed functional relationships.}
  \label{true_func}
\end{figure*}

An example of the true prescribed functional relationships is shown in Figure \ref{true_func}.

\section{Qualitative evaluation: consistency with established ecological knowledge} \label{qualitative}

\begin{figure*}
  \centering
  \includegraphics[width=0.9\textwidth]{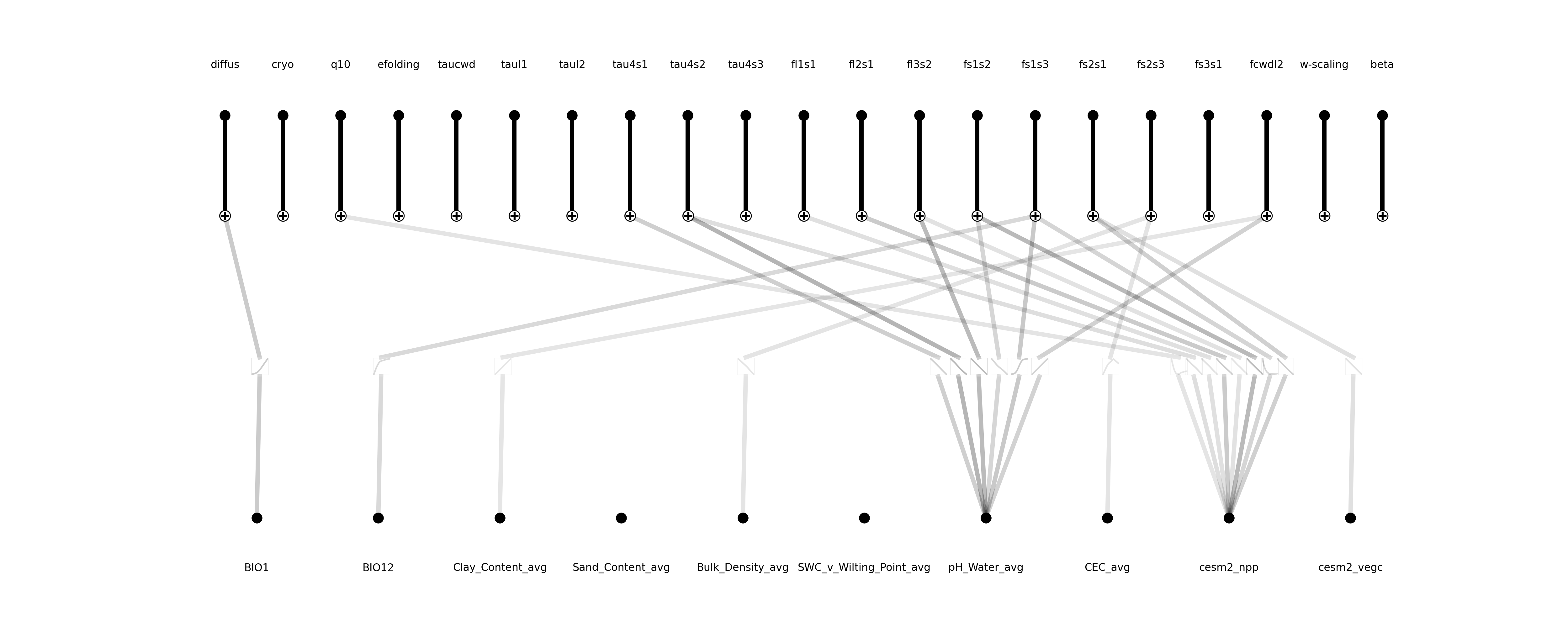}
  \caption{Relationships learned by ScIReN on real data (Table 4).}
  \label{real_relationships}
\end{figure*}
\begin{figure*}
  \centering
  \includegraphics[width=0.9\textwidth]{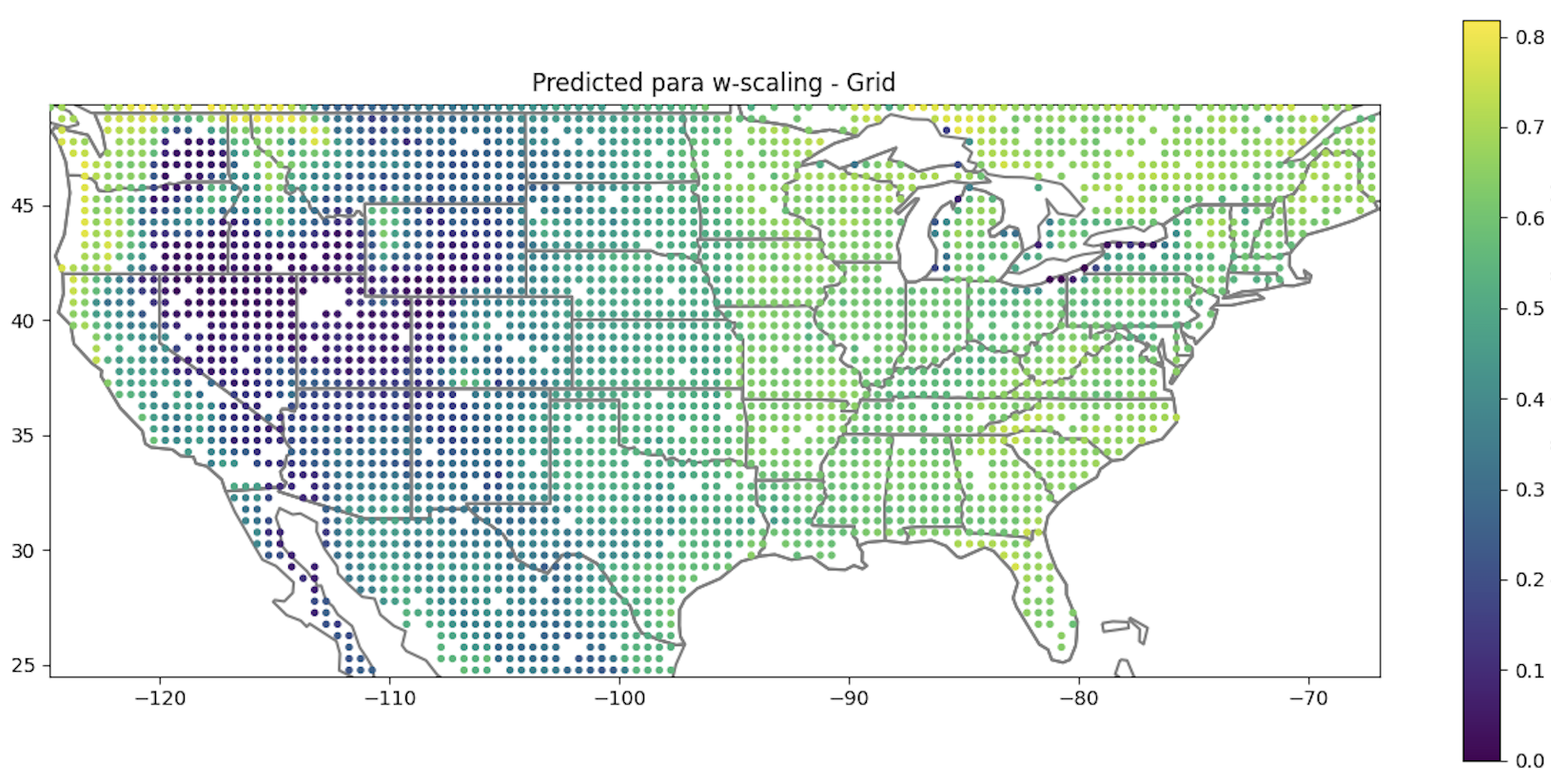}
  \caption{Parameter map (w-scaling) predicted by ScIReN, on real data (Table 4).}
  \label{para_map}
\end{figure*}
For an example of the relationships learned between environmental inputs and biogeochemical relationships on real US data for the CLM5 model, see Figure \ref{real_relationships}. While we do not have ground-truth data to directly evaluate these relationships, the learned relationships are consistent with established ecological knowledge. For example, ScIReN found a positive exponential-like relationship between mean annual temperature (BIO1) and diffusion rate (diffus) in vertical transport, suggesting that higher temperatures will accelerate the vertical movement of organic carbon. Such a relationship agrees well with the conventional understanding that higher temperatures provide more kinetic energy to support faster diffusion \cite{taylor1938temperature}. Meanwhile, we found spreading negative relationships between fresh plant carbon input (NPP) and parameters related to carbon transfer efficiencies (f\_ij) and SOC substrate baseline turnover times (tau\_i). These emerging functional relationships support a positive long-term priming effect at the continental scale, where higher rates of plant carbon input will likely lead to accelerated SOC decomposition (lower tau\_i) and eventually less SOC accrual (lower f\_ij) \cite{kuzyakov2010priming}.

ScIReN can be used to infer the spatial patterns of latent biogeochemical parameters across the world. An example is shown in Figure \ref{para_map}, which displays predicted values of ``w-scaling'', which is a scalar used in CLM5 to indicate the influence of soil water condition on SOC decomposition. A smaller value of w-scaling suggests stronger soil water stress for soil microbes to decompose SOC in their metabolisms, leading to a lower decomposition rate when all other conditions are kept the same \cite{tao2024convergence}. We found that ScIReN successfully learnt the soil water stress across the conterminous United States, showing lower values in arid regions, such as the Interior West, and higher values in relatively more humid regions, such as the East and West Coasts.

\section{Soil carbon cycle: Hyperparameters}

For soil carbon cycle: for neural networks, we used a 3-layer MLP with 128 hidden units, residual connection, batch normalization, and leaky ReLU activation. We set the temperature $\tau=1$. For KANs, we used cubic splines, 30 grid points, and a margin of 2 times the input range; we also used 128 hidden units for the 2-layer KAN (probably too many). We set $\lambda_{param}=1000$ if using hardsigmoid (linear constraint).

We tuned the learning rate from $[10^{-4}, 10^{-3}, 10^{-2}, 10^{-1}]$, weight decay from $[0, 10^{-4}, 10^{-3}]$, $\lambda_{entropy},\lambda_{L1}$ from $[0.1, 1, 10]$, and $\lambda_{smooth}$ from $[10, 100, 1000]$. The weights are higher than for ecosystem respiration because the magnitude of the supervised loss is 100-1000 times larger.

Table \ref{hyperparam3} contains hyperparameters used for Table 3. These were trained using a single CPU.

\begin{table*}
  \caption{Hyperparameters for Table 3.}
  \label{hyperparam3}
  \centering
  \begin{tabular}{lcccc|c}
    \toprule
    \textbf{Method} & LR & Weight decay & $\lambda_{entropy}, \lambda_{L1}$ & $\lambda_{smooth}$ & Runtime \\ \hline
    Pure-NN & $10^{-2}$ & $10^{-4}$ & - & - &  15min \\ 
    Blackbox-Hybrid, nonlinear constraint & $10^{-2}$ & $0$ & - & - & 180min \\
    Blackbox-Hybrid, linear constraint & $10^{-2}$ & $10^{-4}$ & - & - & 180min \\
    Linear-Hybrid, linear constraint & $0.1$ & $10^{-4}$ & - & - & 120min \\
    ScIReN, linear constraint (1-layer KAN) & $10^{-2}$ & 0 & 10 & 10 & 190min \\ \hline
  \end{tabular}
\end{table*}

Table \ref{hyperparam4} contains hyperparameters used for Table 4. These were trained using PyTorch DistributedDataParallel on 8 CPUs.

\begin{table*}
  \caption{Hyperparameters for Table 4.}
  \label{hyperparam4}
  \centering
  \begin{tabular}{lcccc|c}
    \toprule
    \textbf{Method} & LR & Weight decay & $\lambda_{entropy}, \lambda_{L1}$ & $\lambda_{smooth}$ & Runtime \\ \hline
    Pure-NN & 0.1 & 0 & - & - & 2min \\ 
    Blackbox-Hybrid, nonlinear constraint & $10^{-3}$ & 0 & - & - & 25min \\
    Blackbox-Hybrid, linear constraint & $10^{-2}$ & $10^{-4}$ & - & - & 25min \\
    Linear-Hybrid, linear constraint & $10^{-2}$ & 0 & - & - & 40min \\
    ScIReN, linear constraint (1-layer KAN) & $10^{-2}$ & 0 & 1 & 100 & 40min \\
    ScIReN, linear constraint (2-layer KAN) & $10^{-2}$ & 0 & 1 & 100 & 80min \\ \hline
  \end{tabular}
\end{table*}

\section{Computational Resources}

For ecosystem respiration, experiments were run on a cluster, using a single NVIDIA V100 GPU. For CLM5, experiments were run on a cluster using 1 or 8 CPUs (no GPUs). (For the synthetic experiment in Table 3, using 1 CPU performed better than 8 CPUs, possibly because distributed training with 8 CPUs made it hard for each worker's gradient updates to agree on which conections to choose.) For the real label experiments used 8 CPUs with distributed training. Some runs were also performed on a laptop with M3 chip. 

We did not log the total amount of compute required for all failed experiments. The total amount of cluster usage during this period was about $17,000$ CPU-hours; this included a small amount of time spent on other projects, but excluded runs on a local laptop, which may be roughly equivalent.

\section{Ablations} \label{ablations}

To justify the components of ScIReN, in Table \ref{ablation} we show the results of removing each loss term, on Table 2 (ecosystem respiration, nonlinear $R_b$, 2-layer KAN)

\begin{table*}
  \caption{Ecosystem respiration, nonlinear $R_b$. Ablations.}
  \label{ablation}
  \centering
  \begin{tabular}{lccc}
    \toprule
    \textbf{Method} & $R^2$ (observed, $\uparrow$) & $R^2$ (latent, $\uparrow$) & KL, functional relationships $(\downarrow)$ \\ \hline
    Original & \textbf{0.9619} & 0.9989 & \textbf{0.1490} \\
    Remove smoothness loss & 0.9600 & 0.9720 & 0.4973 \\
    Remove L1 (absolute deviation) loss & 0.9618 & \textbf{0.9993} & 0.2016\\
    Remove entropy loss & 0.9608 & 0.8754 & 0.6132 \\
    Reduce to 3 grid cells & 0.9558 & 0.9863 & 0.8942 \\ \hline
  \end{tabular}
\end{table*}

Most methods do similarly on the observed variable. On the latent variable, removing L1 (absolute deviation) loss actually did slightly better; the L1 loss and entropy loss are somewhat interchangable as they both aim to shrink the variance explained by most features to zero. In predicting functional relationships, all losses seem quite valuable. Removing the entropy loss caused a large drop in performance, as the model would pay too much attention to correlated features instead of the true causal one. Finally, reducing to 3 grid cells makes it impossible for the network to learn sharp transitions like the absolute value function.

\ifreproStandalone
\end{document}
\fi

\end{document}